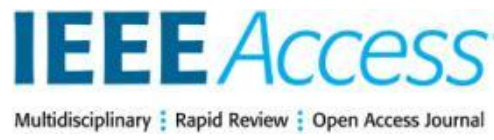

# CSC-Unet: A Novel Convolutional Sparse Coding Strategy based Neural Network for Semantic Segmentation

**HAITONG TANG[1], SHUANG HE[1], MENGDUO YANG[2], XIA LU[3], QIN YU[4], KAIYUE LIU[1], HONGJIE YAN[5], AND NIZHUAN WANG[1, 6, 7]**

[1]School of Geomatics and Marine Information, Jiangsu Ocean University, Lianyungang 222005, China
[2]School of Information Technology, Suzhou Institute of Trade & Commerce, Suzhou, China 215009
[3]School of Geography Science and Geomatics Engineering, Suzhou University of Science and Technology, Suzhou 215000, China
[4]School of Computer Engineering, Jiangsu Ocean University, Lianyungang 222005, China
[5]Department of Neurology, Affiliated Lianyungang Hospital of Xuzhou Medical University, Lianyungang 222002, China
[6]School of Biomedical Engineering, ShanghaiTech University, Shanghai 201210, China
[7]Department of Chinese and Bilingual Studies, The Hong Kong Polytechnic University, Hung Hom, Kowloon, Hong Kong Special Administrative Region, China

Corresponding author: Hongjie Yan (yanhjns@gmail.com) or Nizhuan Wang (wangnizhuan1120@gmail.com).

This work was supported by National Natural Science Foundation of China (No. 41506106, 61701318, 82001160), Natural Science Research Project of Jiangsu Higher Education Institutions (No. 20KJB520014), Project of Huaguoshan Mountain Talent Plan - Doctors for Innovation and Entrepreneurship and Jiangsu Province Graduate Research and Practice Innovation Program Project (No. SY202144X).

**ABSTRACT** It is still a challenging task to perform the semantic segmentation with high accuracy due to the complexity of real picture scenes. Many semantic segmentation methods based on traditional deep learning insufficiently captured the semantic and appearance information of images, which put limit on their generality and robustness for various application scenes. Thus, in this paper, we proposed a novel strategy that reformulated the popularly used convolution operation to multi-layer convolutional sparse coding block in semantic segmentation method to ease the aforementioned deficiency. To prove the effectiveness of our idea, we chose the widely used U-Net model for the demonstration purpose, and we designed CSC-Unet model series based on U-Net. Through extensive analysis and experiments, we provided credible evidence showing that the multi-layer convolutional sparse coding block enables semantic segmentation model to converge faster, extract finer semantic and appearance information of images, and improve the ability to recover spatial detail information. The best CSC-Unet model significantly outperforms the results of the original U-Net on three public datasets with different scenarios, i.e., 87.14% vs. 84.71% on DeepCrack dataset, 68.91% vs. 67.09% on Nuclei dataset, and 53.68% vs. 48.82% on CamVid dataset, respectively. In addition, the proposed strategy could be possibly used to significantly improve segmentation performance of any semantic segmentation model that involves convolution operations and the corresponding code is available at https://github.com/NZWANG/CSC-Unet.

**INDEX TERMS** U-Net, Semantic segmentation, Deep learning, Convolution operation, Convolutional sparse coding (CSC).

## I. INTRODUCTION

In reality, the increasing application scenarios require inferring relevant knowledge or semantics from images, as a result, the importance of semantic segmentation for scene understanding is gradually increasing. Semantic segmentation gives us more detailed understanding of images than image classification [1]–[5] or object detection [6]–[14]. This understanding is crucial in many different domains such as autonomous driving [15]–[17], robotics [18]–[20], image search engines [21]–[23], etc. Recently, many semantic segmentation methods have emerged. For example, fully convolutional networks (FCN) [24], at an end to end form, has firstly implemented the pixel-wise prediction task based on convolution operation, achieving relatively better results in natural scene image segmentation.

SegNet [25] makes the model more efficient than FCN by introducing more skip architectures and max-pooling indexes. PSPNet [26] used dilated convolution and pyramid pooling to improve SegNet. U-Net [27] was proposed in the 2015 ISBI competition, which consists of contracting and symmetrically expanding sub-networks to form a U-shaped architecture. This model was originally designed to solve biomedical image segmentation. Since it requires a small number of training samples to achieve good segmentation results. There are also many variants of the U-Net. For example, Unet++ [28] concatenates U-Net models with different layers that share weights in encoding blocks and features from different layers are also fused through skip connections to achieve better segmentation results. Based on U-Net, FPN [29] and Resnet [5], the U2-Net [30] model is proposed, which achieves surprising results on saliency detection tasks with good real-time performance. Other excellent semantic segmentation models include DeepLab series. For example, Deeplabv1 [31] uses VGG16 [3] with atrous convolutions as the backbone, and adds conditional random fields (CRFs) in post-processing to further improve segmentation performance. Deeplabv2 [32] replaces the backbone in Deeplabv1 with ResNet and proposes atrous spatial pyramid pooling (ASPP) for multi-scale segmentation. Deeplabv3 [33] increases the depth of the backbone without CRFs. It also replaced the convolution with atrous rate of 24 in the ASPP with a 1×1 convolution and added average pooling and batch normalization layers [34]. Deeplabv3+ [35] designs Deeplabv3-based encode-decode models and modifies Xception [36] as the backbone.

All the above semantic segmentation models are based on convolution operations, which have strong feature representation capabilities to extract semantic (global) and appearance (local) information of images. In fact, for the segmentation task of complex image, it is usually limited by the semantic and appearance information extracted from the shallow convolution layers. As a rule, they mostly choose to deepen the network layers so that the semantic segmentation network can better capture the semantic and appearance information of the images to improve the segmentation performance. However, if the network keeps deepening indefinitely, there is a tremendous challenge for both the computational power and the optimizer. Therefore, the main motivation of this study is to address the problem of insufficient feature extraction of convolution operation at the root by optimizing instead of deepening them.

In this paper, we proposed a novel strategy in semantic segmentation model which reformulated convolution operation to multi-layer convolutional sparse coding (ML-CSC) block. Taking the U-Net as an example, we demonstrated the effectiveness and robustness of ML-CSC block strategy in the designed CSC-Unet model series, and it can also be potentially applied to other excellent convolution-based semantic segmentation networks, such as SegNet, U2-Net, etc. Actually, in the Appendix part, we have also implemented the CSC-Unet++, CSC-Unet3+, and CSC-DeepLabv3+ models corresponding to Unet++, Unet3+ [37], and DeepLabv3+, respectively. Benefit from the advantages of ML-CSC block in information representation compared to convolutional operation, we hypothesize that the CSC-Unet model series has the superiorities of better captured semantic and appearance information of original images, better spatial detail information, and better convergence efficiency without increasing the trainable parameters.

As far as we are aware, it is the first work to explore semantic segmentation based on convolutional sparse coding. We hope that our strategy can provide new insights for designing semantic segmentation models. The main contributions of this paper are summarized as follows:

1) We proposed a novel strategy in semantic segmentation networks that used the multi-layer convolutional sparse coding blocks instead of the traditional convolution operations;

2) We extended the ML-CSC block to U-Net as CSC-Unet model series, and further demonstrated the advantages and feasibility through extensive experiments;

3) We explored the impact of the number of unfoldings in the ML-CSC block on the performance of the semantic segmentation model.

The rest of this paper is organized as follows. Section II will give a brief introduction of sparse coding, ML-CSC and block of ML-CSC. In Section III, we will present the design details of CSC-Unet model series. The procedure and results of the experiments are presented in Section IV. Section V is the conclusion and future work that we will carry out.

## II. Review OF MULTI-LAYER CONVOLUTIONAL SPARSE CODING.

In this section, we firstly reviewed the sparse coding, multi-layer convolutional sparse coding and its solution algorithms, then we presented the details of the designed ML-CSC block.

### A. Sparse Coding

Sparse coding represents the signal with few non-zero coefficients as possible and has been used in a wide variety of applications [38]–[43]. In which, the image $\mathbf{y}$ is considered as a linear combination of a set of basis vectors $\mathbf{d}_i$, where most of the coefficients $\gamma_i$ are zero.

$$\mathbf{y} = \mathbf{D\Gamma} = \begin{bmatrix} \mathbf{d_1} & \mathbf{d_2} & \cdots & \mathbf{d}_K \end{bmatrix} \begin{bmatrix} \gamma_1 \\ \gamma_1 \\ \vdots \\ \gamma_K \end{bmatrix} \tag{1}$$

However, when processing image signal, sparse coding first decomposes the whole image into a set of overlapping image blocks and then operates these blocks independently, which leads to that the sparse representation of the image is highly redundant and loss of detailed information during image recovery [44].

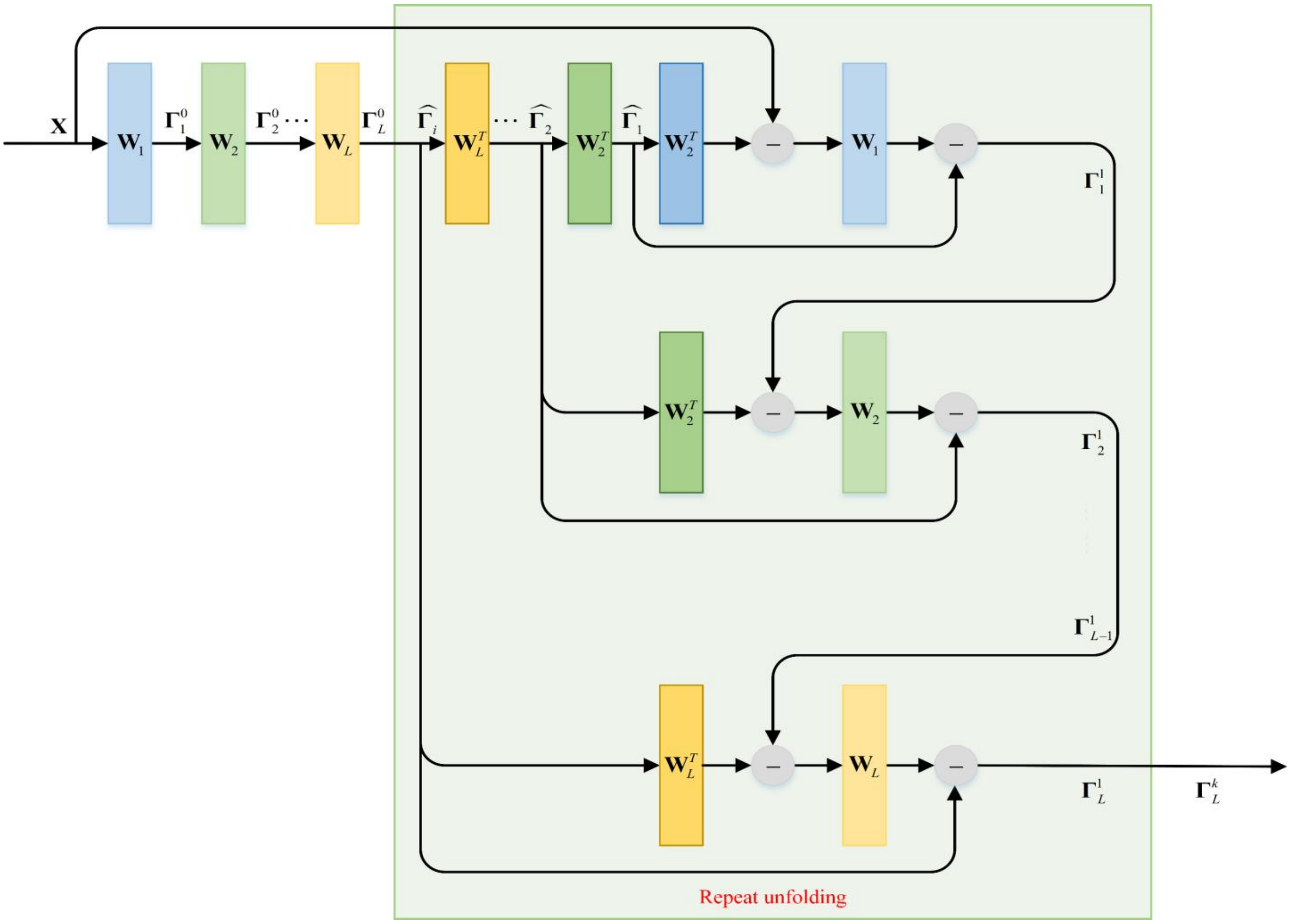

**FIGURE 1.** **The ML-CSC block. $\{\mathbf{W}_i\}_{i=1}^{L}$ denote the convolution operation, $\{\mathbf{W}_i^T\}_{i=1}^{L}$ denote the deconvolution operation, *L* denotes the number of layers, and *k* denotes unfolding number.**

### *B. Multi-Layer Convolutional Sparse Coding*

Assumes that the input image $\mathbf{y}$ satisfies ML-CSC model, it can be denoted as:

$$\begin{aligned} \mathbf{y} &= \mathbf{D}_1\mathbf{\Gamma}_1, \\ \mathbf{\Gamma}_1 &= \mathbf{D}_2\mathbf{\Gamma}_2, \\ &\vdots \\ \mathbf{\Gamma}_{L-1} &= \mathbf{D}_L\mathbf{\Gamma}_L. \end{aligned} \tag{2}$$

where $\{\mathbf{D}_i\}_{i=1}^{L}$ are special dictionaries, each $\mathbf{D}_i$ is a transpose of a convolutional operator matrix $\mathbf{W}_i$ .

$$\mathbf{D}_i = \mathbf{W}_i^T \tag{3}$$

Convolutional sparse coding (CSC) [45] applies convolutional filters to reconstruct the whole image. Since the image is processed as a whole, it bridges the above gap in sparse coding. ML-CSC is an extension of convolutional CSC which sets that the $\{\mathbf{\Gamma}_i\}_{i=1}^{L}$ also satisfy CSC model to form multi-layer representation method about original image.

### *C. The Solver of Multi-Layer Convolutional Sparse Coding Model*

The process of solving the $\{\mathbf{\Gamma}_i\}_{i=1}^{L}$ in ML-CSC model can be formulated as an optimization problem as Equation 4. Where $\hat{\mathbf{\Gamma}}_0 = \mathbf{y}$ , and $\| \ \|_0$ is sparsity regularization term [46]. $\{\lambda_i\}_{i=1}^{L}$ are regularization parameters to balance the sparsity and accuracy of $\{\mathbf{\Gamma}_i\}_{i=1}^{L}$ .

$$\begin{aligned} \hat{\mathbf{\Gamma}}_1 &= \arg\min_{\mathbf{\Gamma}_1} \frac{1}{2}\left\|\hat{\mathbf{\Gamma}}_0 - \mathbf{D}_1\mathbf{\Gamma}_1\right\|_2^2 + \lambda_1\left\|\mathbf{\Gamma}_1\right\|_0, \\ \hat{\mathbf{\Gamma}}_2 &= \arg\min_{\mathbf{\Gamma}_2} \frac{1}{2}\left\|\hat{\mathbf{\Gamma}}_1 - \mathbf{D}_2\mathbf{\Gamma}_2\right\|_2^2 + \lambda_2\left\|\mathbf{\Gamma}_2\right\|_0, \\ &\vdots \\ \hat{\mathbf{\Gamma}}_L &= \arg\min_{\mathbf{\Gamma}_L} \frac{1}{2}\left\|\hat{\mathbf{\Gamma}}_{L-1} - \mathbf{D}_L\mathbf{\Gamma}_L\right\|_2^2 + \lambda_L\left\|\mathbf{\Gamma}_L\right\|_0. \end{aligned} \tag{4}$$

Finding $\{\mathbf{\Gamma}_i\}_{i=1}^{L}$ at once is NP-hard and challenging in computation and concept because of the inclusion of the 0-norm [47]. Emmanuel et al. [48] have shown that the 0-norm can be deflated to 1- norm, turning Equation 4 into a convex optimization problem, as shown in Equation 5.

TABLE I
THE DETAILS OF MODELS

| Architecture \ Model | U-Net | CSC-Unet-Encode | CSC-Unet-Decode | CSC-Unet-All |
|---|---|---|---|---|
| Encoder | [2×Conv2d] × 5 | [ML-CSC block] × 5 | [2×Conv2d] × 5 | [ML-CSC block] × 5 |
| Decoder | [2×Conv2d] × 4 | [2×Conv2d] × 4 | [ML-CSC block] × 4 | [ML-CSC block] × 4 |

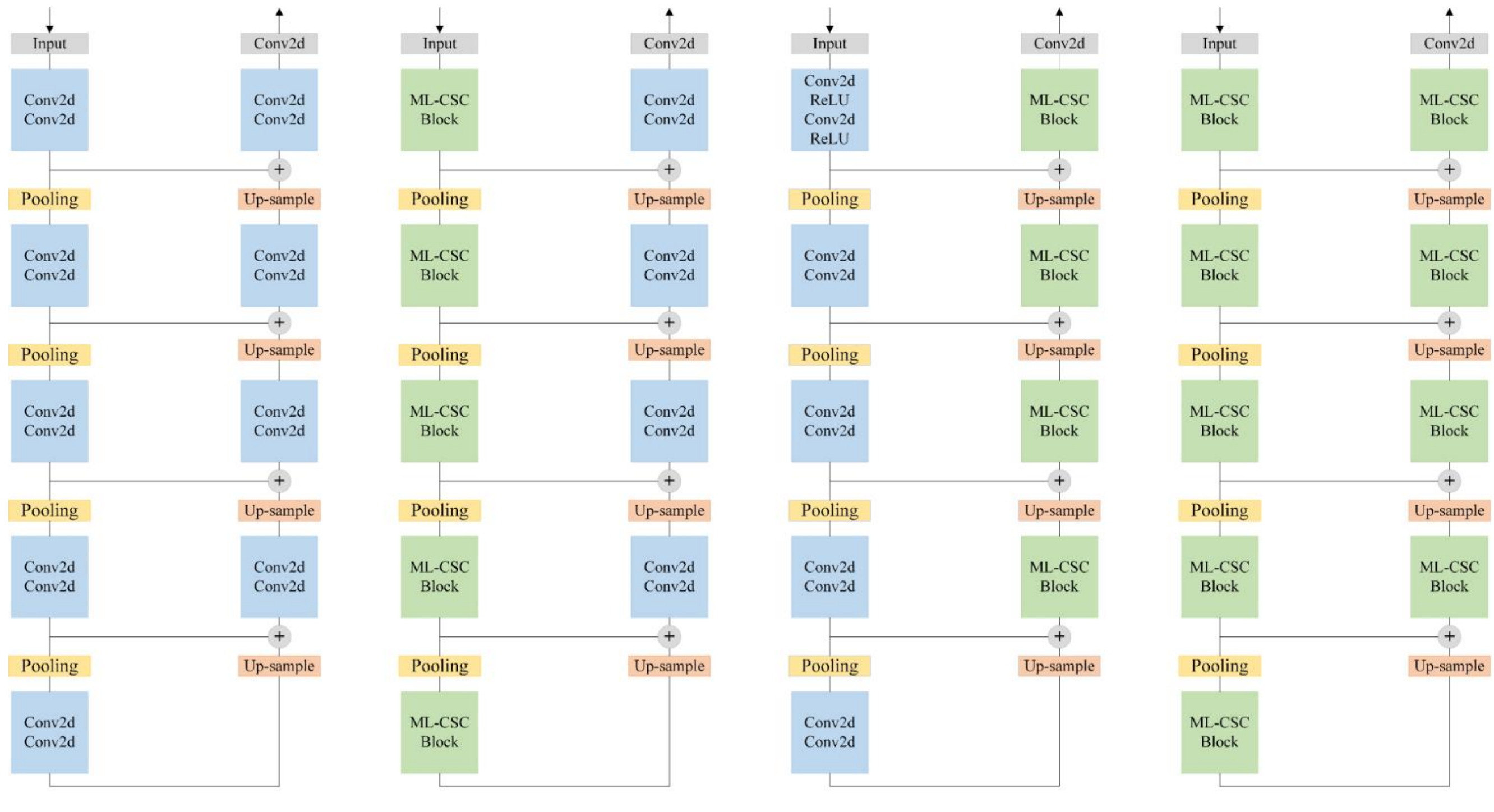

**FIGURE 2. Structures of (a) U-Net, (b) our CSC-Unet-Encode, (c) our CSC-Unet-Decode, and (d) our CSC-Unet-All, respectively.**

$$
\begin{aligned}
\hat{\boldsymbol{\Gamma}}_1 &= \arg\min_{\boldsymbol{\Gamma}_1} \frac{1}{2}\left\|\hat{\boldsymbol{\Gamma}}_0 - \mathbf{D}_1\boldsymbol{\Gamma}_1\right\|_2^2 + \lambda_1\left\|\boldsymbol{\Gamma}_1\right\|_1,\\
\hat{\boldsymbol{\Gamma}}_2 &= \arg\min_{\boldsymbol{\Gamma}_2} \frac{1}{2}\left\|\hat{\boldsymbol{\Gamma}}_1 - \mathbf{D}_2\boldsymbol{\Gamma}_2\right\|_2^2 + \lambda_2\left\|\boldsymbol{\Gamma}_2\right\|_1,\\
&\ \vdots\\
\hat{\boldsymbol{\Gamma}}_L &= \arg\min_{\boldsymbol{\Gamma}_L} \frac{1}{2}\left\|\hat{\boldsymbol{\Gamma}}_{L-1} - \mathbf{D}_L\boldsymbol{\Gamma}_L\right\|_2^2 + \lambda_L\left\|\boldsymbol{\Gamma}_L\right\|_1.
\end{aligned} \tag{5}
$$

### 1) LAYERED THRESHOLDING ALGORITHM

Papyan et al. [49] proposed the layered thresholding algorithm, that use thresholding algorithm [46] to solve the sparse vectors $\{\boldsymbol{\Gamma}_i\}_{i=1}^{L}$ step by step in different layers. The solver can be written as follow:

$$
\begin{aligned}
\hat{\boldsymbol{\Gamma}}_1 &= h_{\theta_1}\left(\mathbf{D}_1^T\mathbf{y}\right),\\
\hat{\boldsymbol{\Gamma}}_2 &= h_{\theta_2}\left(\mathbf{D}_2^T\hat{\boldsymbol{\Gamma}}_1\right),\\
&\ \vdots\\
\hat{\boldsymbol{\Gamma}}_L &= h_{\theta_L}\left(\mathbf{D}_L^T\hat{\boldsymbol{\Gamma}}_{L-1}\right).
\end{aligned} \tag{6}
$$

The soft non-negative threshold operator $\{h_{\theta_i}\}_{i=1}^{L}$ can be viewed as a translation of the activation function rectified linear unit (ReLU) activation function [50] by $\{\theta_i\}_{i=1}^{L}$ units [49]. Combined with Equation 3, the above equation can be written as follow:

$$
\begin{aligned}
\hat{\boldsymbol{\Gamma}}_1 &= \mathrm{ReLU}\left(\mathbf{W}_1\mathbf{y} + \theta_1\right),\\
\hat{\boldsymbol{\Gamma}}_2 &= \mathrm{ReLU}\left(\mathbf{W}_2\hat{\boldsymbol{\Gamma}}_1 + \theta_2\right),\\
&\ \vdots\\
\hat{\boldsymbol{\Gamma}}_L &= \mathrm{ReLU}\left(\mathbf{W}_L\hat{\boldsymbol{\Gamma}}_{L-1} + \theta_L\right).
\end{aligned} \tag{7}
$$

where and $\{\theta_i\}_{i=1}^{L}$ are trainable parameters.

### 2) MULTI-LAYER ITERATIVE SOFT THRESHOLDING ALGORITHM

An attractive approximate solver of Equation 5 is the multi-layer iterative soft thresholding algorithm (ML-ISTA) [51], which uses an iterative soft thresholding algorithm [52] at each layer.

TABLE II
THE DETAILS OF DATASETS USED IN THE EXPERIMENT

| Dataset | Classes | Samples (training) | Samples (validation) | Samples (test) | Samples (total) |
|---|---|---|---|---|---|
| CamVid | 11 | 367 | 101 | 233 | 701 |
| DeepCrack | 2 | 322 | 107 | 108 | 537 |
| Nuclei | 3 | 402 | 134 | 134 | 670 |

TABLE III
RESULT ON DEEPCRACK, NUCLEI AND CAMVID TEST SET OF U-NET AND CSC-UNET-ENCODE MODELS WITH DIFFERENT UNFOLDING NUMBER

| Method | DeepCrack Mean IoU (%) | Nuclei Mean IoU (%) | CamVid Mean IoU (%) |
|---|---|---|---|
| U-Net (CSC-Unet-Encode-0) | 84.71 | 67.09 | 48.82 |
| CSC-Unet-Encode-1 | 86.41 | 67.26 | 52.31 |
| CSC-Unet-Encode-2 | **86.90** | **68.44** | **53.29** |
| CSC-Unet-Encode-3 | 86.20 | 67.71 | 52.43 |

$$\begin{aligned} \hat{\mathbf{\Gamma}}_1 = \hat{\mathbf{\Gamma}}_1^{k+1} &= \mathrm{ReLU}\left(\tilde{\mathbf{\Gamma}}_1 - \mu_1 \mathbf{D}_1^T\left(\mathbf{D}_1 \tilde{\mathbf{\Gamma}}_1 - \hat{\mathbf{\Gamma}}_0\right) + \theta_1\right), \\ \hat{\mathbf{\Gamma}}_2 = \hat{\mathbf{\Gamma}}_2^{k+1} &= \mathrm{ReLU}\left(\tilde{\mathbf{\Gamma}}_2 - \mu_2 \mathbf{D}_2^T\left(\mathbf{D}_2 \tilde{\mathbf{\Gamma}}_2 - \hat{\mathbf{\Gamma}}_1\right) + \theta_2\right), \\ &\vdots \\ \hat{\mathbf{\Gamma}}_L = \hat{\mathbf{\Gamma}}_L^{k+1} &= \mathrm{ReLU}\left(\tilde{\mathbf{\Gamma}}_L - \mu_L \mathbf{D}_L^T\left(\mathbf{D}_L \tilde{\mathbf{\Gamma}}_L - \hat{\mathbf{\Gamma}}_{L-1}\right) + \theta_L\right). \end{aligned} \quad (8)$$

It is well known that the layered thresholding algorithm is the simplest and dumbest pursuit algorithm for ML-CSC [49][53]. This is because it only uses the thresholding algorithm independently at each layer and does not take into account the connection between layers. In contrast, ML-ISTA solves the above problem by constructing $\tilde{\mathbf{\Gamma}}_i = \mathbf{D_i}\mathbf{D_{i+1}}\cdots\mathbf{D_L}\hat{\mathbf{\Gamma}}_L^{\mathrm{k}}$ in each layer which takes full account of the connections between the different layers..

### D. Multi-Layer Convolutional Sparse Coding Block

In the ML-CSC model, we summarize the solving process as shown in Figure 1, and name it as ML-CSC block. $\left\{\mathbf{W}_i\right\}_{i=1}^{L}$ denotes the convolution operation, and $\left\{\mathbf{W}_{i=1}^{T}\right\}_{i=1}^{L}$ denotes the deconvolution operation, where $\left\{\mu_i\right\}_{i=1}^{L}$ and $\left\{\theta_i\right\}_{i=1}^{L}$ are trainable parameters. Solving $\left\{\mathbf{\Gamma}_i\right\}_{i=1}^{L}$ can be seen as the process of extracting features from multi-layer convolution operation as follow:

$$\mathbf{\Gamma}_L = \mathrm{ReLU}\left(\mathbf{W}_L \ \cdots \ \mathrm{ReLU}\left(\mathbf{W}_2 \ \mathrm{ReLU}\left(\mathbf{W}_1 \mathbf{y}\right)\right)\right) \quad (9)$$

When the number of unfoldings is set to 0, the layered thresholding algorithm is performed in the ML-CSC block. When the unfolding number is greater than 0, the ML-ISTA algorithm is executed. From the sparse point of view, due to ML-ISTA algorithm is superior to the layered thresholding algorithm, the ML-CSC block will extract more accurate feature compared with multi-layer convolution operation, which is beneficial to the forward propagation of the neural network, and also can better capture the semantic and appearance information of the image to improve the segmentation performance.

## III. METHOD

### A. U-NET MODEL

The architecture of U-Net model [27] is displayed in Figure 2(a). . For convenience, we use 3×3 convolution layer with padding to keep the same size before and after convolution operation, thus the input size of model is equal to the output size. The up-sampling is performed by 3×3 transposed convolution operation. Batch normalization [34] is after convolution operation and before ReLU activation function.

### B. CSC-UNET MODEL SERIES

In the encode and decode side of the U-Net model, both of which can be seen as a composition of blocks containing two layers of convolution operation, as shown in Table I. To fairly demonstrate our strategy, we set the number of layers in the ML-CSC block to 2 as well. According to the characteristics of encoding and decoding structure in U-Net, we designed the CSC-Unet-model series including CSC-Unet-Encode, CSC-Unet-Decode, and CSC-Unet-All model. The details of the CSC-Unet model series were also shown in Table I.

#### 1) CSC-UNET-ENCODE MODEL

The encoding side of U-Net is used to extract the semantic and appearance information of the input image. In order to explore this kind of ability of ML-CSC block, we replaced the convolution operations of the encoding side of the U-Net with the ML-CSC blocks to form CSC-Unet-Encode model, and the corresponding architecture was  shown in Figure 2(b).

#### 2) CSC-UNET-DECODE MODEL

In the expanding sub-network of U-Net model, it firstly uses skip connection to combine appearance information from the shallow layers and semantic information from the deep layers. Then, the decoding side of U-Net precisely locates the segmentation boundary and gradually recovers the spatial detail information of the image. To explore the ability of ML-CSC block to recover the spatial detail information of image, we introduced the ML-CSC block into the decoding side of U-Net to form CSC-Unet-Encode model, and the corresponding architecture was shown in Figure 2(c).

#### 3) CSC-UNET-ALL MODEL

To explore the impact of ML-CSC block on the overall segmentation performance of U-Net model, we added this block to both the encoding and decoding side of the U-Net to form CSC-Unet-All model, and the corresponding architecture was shown in Figure 2(d).

TABLE IV
RESULT ON DEEPCRACK, NUCLEI AND CAMVID TEST SET OF U-NET AND CSC-UNET-DECODE MODELS WITH DIFFERENT UNFOLDING NUMBER

| **Method** | DeepCrack Mean IoU (%) | Nuclei Mean IoU (%) | CamVid Mean IoU (%) |
|---|---|---|---|
| U-Net (CSC-Unet-Decode-0) | 84.71 | 67.09 | 48.42 |
| CSC-Unet-Decode-1 | **86.29** | **68.20** | 51.49 |
| CSC-Unet-Decode-2 | 85.52 | 67.24 | 52.33 |
| CSC-Unet-Decode-3 | 85.21 | 67.25 | **53.18** |

TABLE V
RESULT ON DEEPCRACK AND NUCLEI TEST SET OF U-NET AND CSC-UNET-ALL MODELS WITH DIFFERENT UNFOLDING NUMBER

| **Method** | DeepCrack | | Nuclei | |
|---|---|---|---|---|
| | Pixel Acc (%) | Mean IoU (%) | Pixel Acc (%) | Mean IoU (%) |
| U-Net (CSC-Unet-All-0-0) | 98.53 | 84.71 | 96.64 | 67.09 |
| CSC-Unet-All-2-1 (best) | **98.74** | **87.14** | **96.81** | **68.91** |
| CSC-Unet-All-1-1 | 98.62 | 86.61 | 96.67 | 67.30 |
| CSC-Unet-All-2-2 | 98.72 | 87.04 | 96.73 | 68.31 |

TABLE VI
RESULTS ON CAMVID TEST SET OF CSC-UNET-ALL MODELS (1) U-NET(CSC-UNET-ALL-0-0), (2) CSC-UNET-ALL-2-1, (3) CSC-UNET-ALL-1-1 AND (4) CSC-UNET-ALL-2-2

| Model | Sky | Building | Pole | Road | Sidewalk | Tree | Sign | Fence | Car | Pedestrian | Building | Class avg. (%) | Mean IOU (%) |
|---|---|---|---|---|---|---|---|---|---|---|---|---|---|
| (1) | **96.6** | 75.9 | 27.7 | 96.4 | 75.1 | 81.2 | **50.1** | 16.3 | **87.0** | 43.6 | 10.9 | 60.1 | 48.82 |
| (2) | 96.1 | 84.6 | 33.4 | 97.6 | 85.2 | 81.2 | 45.6 | **18.4** | 86.0 | 54.8 | 12.5 | 63.2 | 53.56 |
| (3) | 95.9 | 80.5 | 33.6 | **97.8** | **85.8** | 82.8 | 41.8 | 15.2 | 85.5 | **60.6** | **14.5** | 63.1 | 52.76 |
| (4) | 95.5 | **87.5** | **36.6** | 97.6 | 83.3 | **82.9** | 38.4 | 16.5 | 86.0 | 59.6 | 12.2 | **63.3** | **53.68** |

## IV. EXPERIMENT AND ANALYSIS

In this study, the computing platform are Ubuntu 18.04.5 LTS 64-bit OS, 32G RAM, and Nvidia GeForce GTX 1080 Ti GPU with 11 GB memory. The deep learning framework is based on PyTorch [54] and the program language is Python.

### A. DATASETS

For perform a fair evaluation, we select different scenarios datasets for testing to obtain evaluation metrics of comparison models. CamVid dataset [55] is one of the first datasets used for autonomous driving. It is assembled from 5 video sequences taken by the on-board camera from the driver's perspective. DeepCrack dataset [56] is a public benchmark dataset containing cracks at multiple scales and scenarios to evaluate crack detection systems. Nuclei dataset[1] is the dataset in the 2018 Kaggle Data Science Bowl which is acquired under a variety of conditions and variations in the cell type, magnification, and imaging modality. The details of the datasets were shown in Table II.

### B. THE SETTING OF TRAINING PARAMETERS

The number of all epochs were empirically set to 200 in this experiment. To improve the generalization ability of the model, before each epoch, we randomly disrupted the training data to make it more consistent with the sample distribution under natural conditions. The batch size was set to 4. The loss function was negative log-likelihood, and the input parameters were activated by the log-SoftMax function. The model used Adam [57] algorithm as the optimizer, each 50 epochs, and the learning rate dropped by half. In the CamVid dataset. The initial learning rate was set to $10^{-4}$, for the DeepCrack and Nuclei, the initial learning rate was set to $10^{-5}$, respectively.

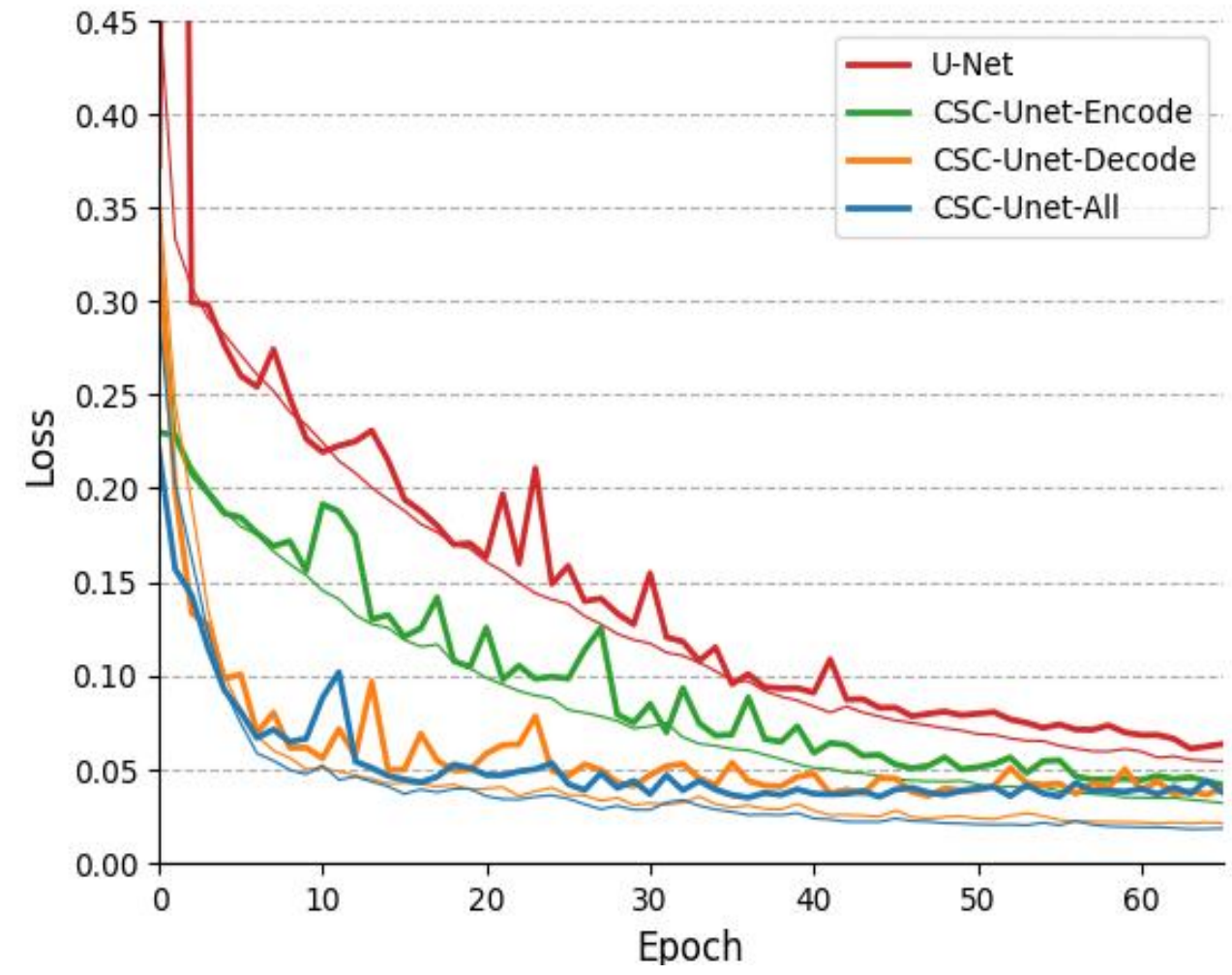

**FIGURE 3. Training and validation on DeepCrack. Fine curves indicate the loss of training, and thick curves indicate the loss of validation. Unfolding number is uniformly set to 2 and the trainable parameters are same for all models.**

[1] https://www.kaggle.com/c/data-science-bowl-2018/overview

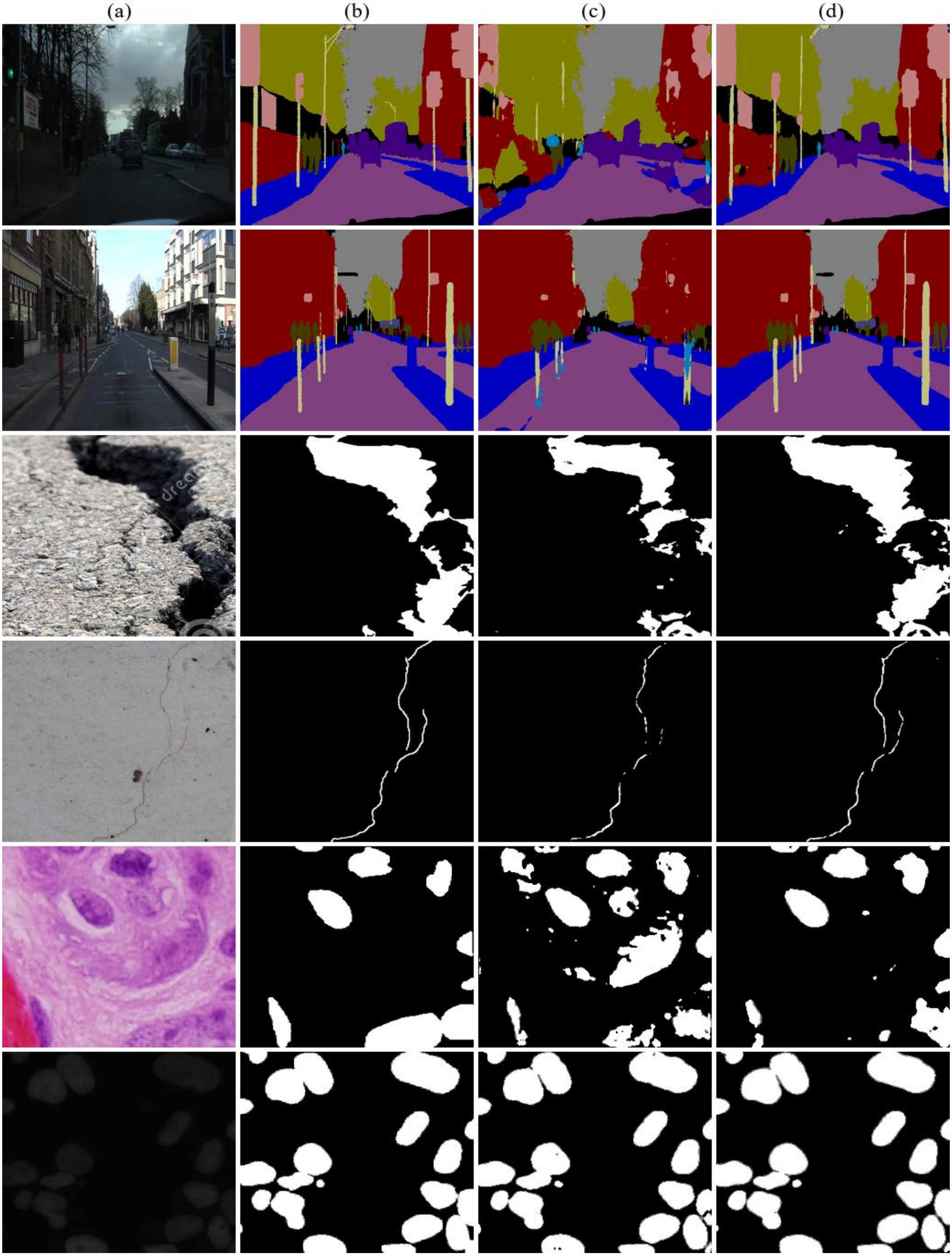

**FIGURE 4.** **Examples of semantic segmentation results on CamVid, DeepCrack, and Nuclei test set. (a) Input images, (b) Ground truths, (c) Results of U-Net, and (d) Results of CSC-Unet-All (best).**

### C. EXPERIMENT AND ANALYSIS

#### 1) THE SPEED OF MODEL CONVERGENCE

We investigated the effect of ML-CSC blocks on the convergence speed of segmentation models. On the DeepCrack dataset, we compared the training and validation loss of CSC-Unet model series in the training phase, and the results were shown in Figure 3. We found that ML-CSC blocks can accelerate the convergence of the semantic segmentation model and reduce the loss value. Adding ML-CSC blocks at the decoding side of the U-Net model converged faster than adding them at the encoding side, and the model converged fastest when ML-CSC blocks were added at both sides of U-Net.

#### 2) THE EXTRACTION OF SEMANTIC AND APPEARANCE INFORMATION

We have assessed the influence of ML-CSC block on CSC-Unet model series compared with U-Net to extract semantic and appearance information, and the results were shown in Table III, where the number after the model indicated the unfolding number of ML-CSC block. When the number of unfoldings was zero, the CSC-Unet-Encode was equivalent to the U-Net model. The results showed that CSC-Unet-Encode models outperformed U-Net model on all three datasets when the number of unfoldings was greater than 0. This indicated that the ML-CSC block can indeed improve the ability of the semantic segmentation model to capture the semantic and appearance information of the image. Furthermore, we found that it was not always true that the larger number of unfoldings of the ML-CSC block implied better performance.

Semantic segmentation model first extracts feature information at the encoding side, and then based on feature information, the model gradually recovers the spatial detail information of the image at the decoding side. As the number of unfoldings increases, the feature information conveys in the model becomes sparser, which is not beneficial for the recovery process at the decoding side. Therefore, we should find a balance point between the extraction of feature information and the recovery process. According to Table III, we inferred that the balance point was reached when the unfolding number was two among most datasets. For the convenience of performance demonstration, in our all experiments, the maximum number of unfoldings was set to three.

#### 3) THE ABILITY TO RECOVER SPATIAL DETAIL INFORMATION

Next, we explored the ability of ML-CSC block to recover spatial detail information of image at the decoding side and the results were shown in Table IV. CSC-Unet-Decode models with unfolding number greater than 0 on different datasets were better than U-Net, which implied that ML-CSC block improved the recovery of spatial detail information compared to the convolution operation. The best unfolding was 1 in DeepCrack and Nuclei, and in CamVid was 3, respectively. We speculated that phenomenon was probably related to the complexity of the image, where the categories of DeepCrack and Nuclei were relatively few and the case of one unfolding was enough, but CamVid was relatively more complex and required a higher number of unfoldings.

#### 4) THE OVERALL IMPROVEMENT OF MODEL PERFORMANCE

We first introduced the nomenclature of CSC-Unet-All-a-b, where a denoted the unfolding number at the encoding side and b denoted the number of unfoldings at the decoding side. For example, CSC-Unet-All-0-0 was equivalent to U-Net model, CSC-Unet-All-a-0 was represented as CSC-Unet-Encode-a, and CSC-Uner-All-0-b can be represented as CSC-Unet-Decode-b. Through Table III and IV, we found that on the DeepCrack dataset the CSC-Unet-Encode-2 captured more semantic and appearance information, and CSC-Unet-Decode-1 maximized the ability of the model to recover spatially detailed information. Thus, we argued that CSC-Unet-All-2-1 could maximize the segmentation performance of the U-Net model. Similarly, on Nuclei and CamVid, CSC-Unet-All-2-1 and CSC- Unet-All-2-3 should achieve the best segmentation performance. However, due to the GPU memory size limitation (11 GB), if we use the CSC-Unet-All-2-3 model on the CamVid dataset, the batch size needs to be halved, or we can use GPU parallelism to maintain the batch size. This is something that we do not expect to see. We want to compare the performance of the models under the same conditions; thus, we used CSC-Unet-All-2-2 instead of CSC-Unet-All-2-3. We also set CSC-Unet-All-0-0 (U-Net) and CSC-Unet-All-1-1 for performance comparison. The results on the three datasets were shown in Table V and VI. Compared to U-Net we have improved by 2.43% (from 84.71% to 87.14% in DeepCrack), 1.82% (from 67.09% to 68.91% in Nuclei), and 4.86% (from 48.82% to 53.68% in CamVid) on three datasets in terms of Mean Intersection over Union (MIoU), respectively. To better illustrate the results, the visualizations on the three datasets were shown in Figure 4.

## V. CONCLUSION AND DISCUSSION

In this paper, we proposed a novel strategy, which used ML-CSC block instead of convolutional operation to improve the performance of semantic segmentation model. This strategy could possibly apply to any semantic segmentation model that involved convolutional operation. We used the U-Net model as an example to validate this strategy and designed CSC-Unet model series. We found that using ML-CSC blocks instead of convolution operations could accelerate the convergence of the semantic segmentation model, improve the ability of the model to capture the semantic and appearance information of the image, and improve the ability to recover spatial detail information. We concluded that ML-CSC block was a better operation compared to convolutional operation in semantic segmentation.

The current CSC-Unet models have achieved significant improvement in segmentation performance compared with the original U-Net model. However, they still face great challenges. Therefore, in the follow-up study, we will consider the following aspects.

1) To fairly demonstrate our strategy, the added ML-CSC block to the U-Net model was a two-layer convolutional sparse coding model. We speculate that more layers will be beneficial to improve the performance of semantic segmentation. Thus, in the next study, we will design a new type of semantic segmentation model based on multi-layer global convolutional sparse coding block.

2) We found that there is a balance between the extraction of feature information and the recovery process. Thus, how to find the best balance between all the unfolding number in separate encoder and decoder part will be one focus of our future research. In addition, the presence of the unfolding number in the solution algorithm slightly increases the model's computation cost, as shown in Table VII. So, the solution algorithm without unfolding number will be designed in our subsequent study.

3 ) The ML-CSC block can be extended to not only the field of semantic segmentation, but also possibly to other fields such as object detection [12]–[14], generative adversarial networks (GAN) [58], natural language processing (NLP) [59], etc.

TABLE VII
THE COMPUTATIONAL COST OF U-NET AND CSC-UNET-MODEL SERIES WITH DIFFERENT UNFOLDING NUMBER

| **Model** | Training Memory Usage (GB) | | |
|---|---|---|---|
| | CamVid | DeepCrack | Nuclei |
| Unet | 5.8 | 4.9 | 4.9 |
| CSC-Unet-Encode-1 | 7.0 | 5.8 | 5.8 |
| CSC-Unet-Encode-2 | 8.2 | 6.7 | 6.6 |
| CSC-Unet-Encode-3 | 9.3 | 7.6 | 7.6 |
| CSC-Unet-Decode-1 | 7.3 | 6.1 | 6.2 |
| CSC-Unet-Decode-2 | 8.8 | 7.1 | 7.2 |
| CSC-Unet-Decode-3 | 10.2 | 8.1 | 8.3 |
| CSC-Unet-All-1-1 | 8.5 | 6.9 | 7.1 |
| CSC-Unet-All-2-1 | 9.6 | 7.8 | 8.0 |
| CSC-Unet-All-2-2 | 11.1 | 8.9 | 9.0 |

## APPENDIX

The results of improving Unet++, Unet3+, and deeplabv3+ are shown in TABLE VIII, with the same training strategy and input model sizes as CSC-Unet, and the code of CSC-Unet++, CSC-Unet+++, and CSC-DeepLabv3+ is also available at https://github.com/NZWANG/ CSC-Unet.

TABLE VIII
THE MODEL PERFORMANCE AND COMPUTATIONAL COST OF UNET++, UNET+++, DEEPLABV3+ AND THEIR CORRESPONDING IMPROVED VERSIONS

| **Model** | Unfoldings | Mean IoU (%) | | | Training Memory Usage (GB) | | |
|---|---|---|---|---|---|---|---|
| | | CamVid | DeepCrack | Nuclei | CamVid | DeepCrack | Nuclei |
| Unet++ | - | 58.04 | 98.75 | 88.50 | 5.2 | 4.1 | 7.5 |
| CSC-Unet++ | 1 | 65.44 | **98.98** | **89.91** | 9.0 | 6.9 | 13.3 |
| CSC-Unet++ | 2 | **65.87** | 98.96 | 89.74 | 12.7 | 9.7 | 18.0 |
| Unet3+ | - | 60.09 | 98.92 | 89.46 | 14.6 | 11.3 | 16.1 |
| CSC- Unet3+ | 1 | **66.89** | 99.06 | 90.06 | 15.9 | 12.2 | 18.1 |
| CSC- Unet3+ | 2 | 66.17 | **99.07** | **90.10** | 16.9 | 13.0 | 19.6 |
| DeepLabv3+ | - | 58.98 | 98.81 | 86.92 | 8.5 | 6.5 | 6.4 |
| CSC- DeepLabv3+ | 1 | 63.28 | 98.89 | 87.37 | 13.6 | 10.3 | 10.0 |
| CSC- DeepLabv3+ | 2 | **64.69** | **99.12** | **88.01** | 17.6 | 13.1 | 12.0 |

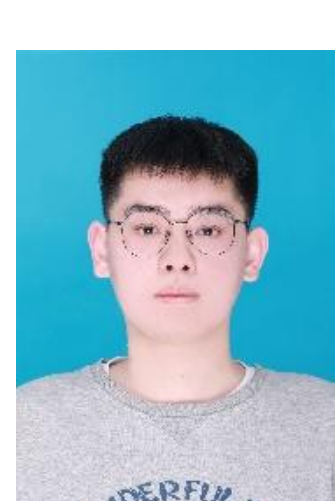

**HAITONG TANG** graduated from Jiangsu Ocean University in 2022 with a master's degree in Surveying and Mapping Engineering. He is currently employed as a Deep Learning Algorithm Engineer at the Jiangsu Dongying Intelligent Engineering Research Institute affiliated with Southeast University. His research focuses on deep learning, computer vision, and convolutional sparse coding.

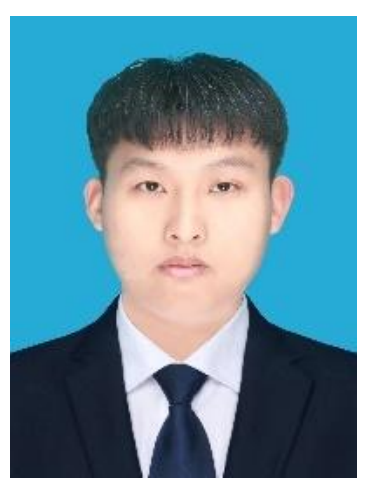

**SHUANG HE** received the B.S. and M.S. degrees from Jiangsu Ocean University, Lianyungang, China, in 2018 and 2022, respectively. He is currently pursuing his doctorate degree in the International Institute for Earth System Science with Nanjing University. His research interests include deep learning, remote sensing image processing, hyperspectral vegetation quantitative analysis.

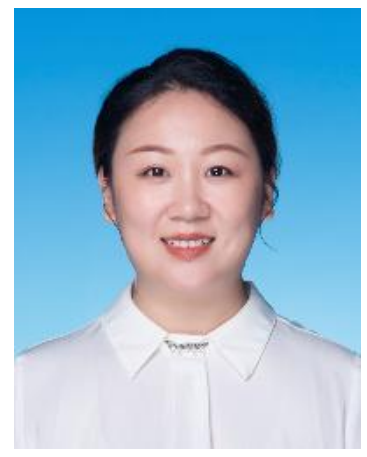

**MENGDUO YANG** received the B.S. degree in computer science and technology from Soochow University, Suzhou, China, in 2011. She also received the Ph.D. degree in software engineering from Soochow University, in December 2016.

Her research interests include medical image analysis and few-shot learning.

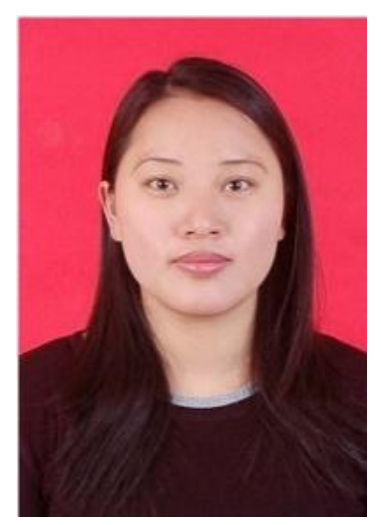

**XIA LU** received the B.S. degree in Exploration Engineering from Jilin University, Jilin, China, in 1997, the M.S. degree in Cartography and Geographic Information Engineering from the China University of Geoscience (Beijing), Beijing, China, in 2004, and the Ph.D. degree in Photogrammetry and Remote Sensing from China University of Mining (Beijing), Beijing, China, in 2008.

She is currently a Professor with School of Geography Science and Geomatics Engineering, Suzhou University of Science and Technology. Her research interests include three-dimensional monitoring and evaluation of marine environment, quantitative remote sensing of coastal wetland ecology.

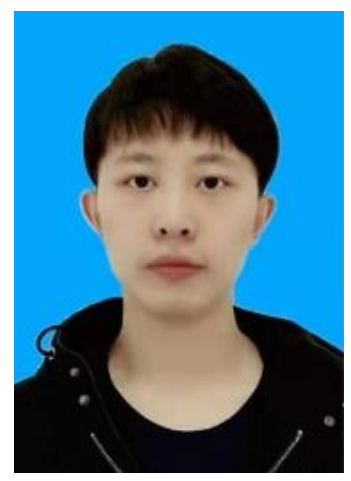

**QIN YU** received his master's degree in Pattern Recognition and Intelligent Systems from Jiangsu Ocean University in 2022. He is currently pursuing his doctorate degree in Biomedical Engineering at Shenzhen University. His research interests include neuroscience, image analysis, and emotions.

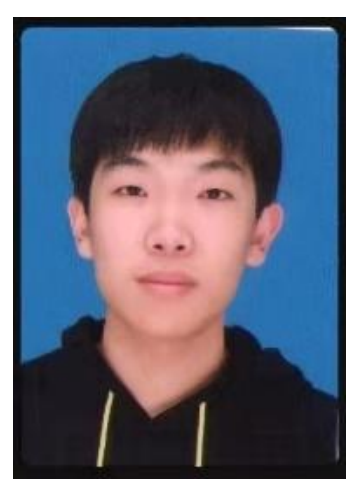

**KAIYUE LIU** received the B.S. and M.S. degrees graduated from Shanxi University of Engineering and Technology and Jiangsu Ocean University in 2018 and 2023, respectively. He is currently pursuing his doctorate degree at Shenzhen University. His research interests include the field of remote sensing, machine learning, object detection and ROS systems.

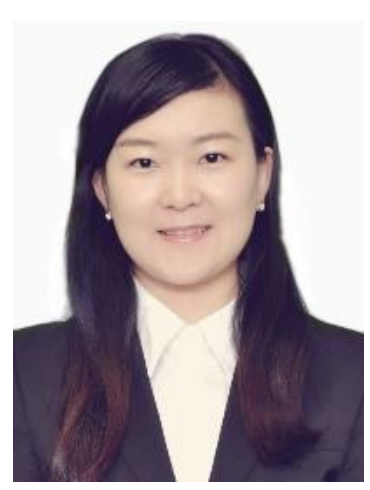

**HONGJIE YAN** received the Ph.D. degree in clinical medicine from Shimane University, Japan, in 2016. She is an attending physician of Department of Neurology, Affiliated Lianyungang Hospital of Xuzhou Medical University, Lianyungang, China.

Her research interests include AI-based diagnosis of neurological disorders, neuroimaging and aging, neurodegenerative diseases.

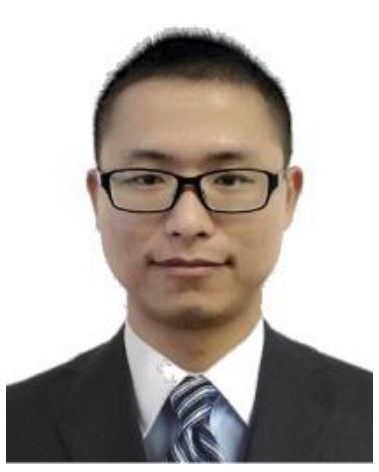

**NIZHUAN WANG** received the B.S. degree in computer science and technology from Heilongjiang University, Harbin, China, in 2010. He received the M.S. degree in computer application technology from Shanghai Maritime University, Shanghai, China, in 2012, where he also received the Ph.D. degree in communications, information engineering, and control, in January 2016.

Currently, he is a Research Assistant Professor with Department of Chinese and Bilingual Studies, The Hong Kong Polytechnic University. Before that, he has sequentially worked at Shenzhen University (as an Assistant Professor), Jiangsu Ocean University (as a full Professor), and ShanghaiTech University (as an Associate Researcher) from 2016 to 2024. His research interests include intelligent computation, neural engineering, medical imaging and hyperspectral imaging, and he has published more than 60 scientific papers on many journals, i.e., Human Brain Mapping, Journal of Neuroscience Methods, Magnetic Resonance Imaging, IEEE Journal of Biomedical and Health Informatics, etc.